\documentclass{sig-alternate}

\usepackage{graphicx}

\begin{document}
\conferenceinfo{AAMAS'05,} {July 25-29, 2005, Utrecht, Netherlands.}
\CopyrightYear{2005}
\crdata{1-59593-094-9/05/0007}

\title{Polymorphic Self-* Agents for Stigmergic Fault Mitigation in Large-Scale Real-Time Embedded Systems}
\date{}
\author{
\alignauthor Derek Messie\\
Jae C. Oh\\
\affaddr{Department of Electrical Engineering and Computer Science}\\
\affaddr{Syracuse University}\\
\affaddr{Syracuse, NY 13244 USA}\\
\email{dsmessie@syr.edu, jcoh@ecs.syr.edu}
\vspace{.2in}
}
\maketitle
\bibliographystyle{abbrv}

\begin{abstract}
Organization and coordination of agents within large-scale, complex, distributed environments is one of the primary challenges in the field of multi-agent systems.  A lot of interest has surfaced recently around self-* (self-organizing, self-managing, self-optimizing, self-protecting) agents.  This paper presents polymorphic self-* agents that evolve a core set of roles and behavior based on environmental cues.  The agents adapt these roles based on the changing demands of the environment, and are directly implementable in computer systems applications.  The design combines strategies from game theory, stigmergy, and other biologically inspired models to address fault mitigation in large-scale, real-time, distributed systems.  The agents are embedded within the individual digital signal processors of BTeV, a High Energy Physics experiment consisting of 2500 such processors.  Results obtained using a SWARM simulation of the BTeV environment demonstrate the polymorphic character of the agents, and show how this design exceeds performance and reliability metrics obtained from comparable centralized, and even traditional decentralized approaches.
\end{abstract}




\section{Introduction}
In the field of multi-agent systems, a lot of attention has been focused lately on investigating various architectures and methodologies that promote effective organization and coordination within large-scale, complex, distributed systems \cite{de:sensorcoord99}\cite{fb:lifecyclecoord02}.  Specifically, the interest is in developing approaches that can be implemented within multi-agent systems to produce some desirable emergent behavior that coordinates individual actors in a system competing for resources such as bandwidth, computing power, and data.

Agent methodologies that exhibit self-* (self-organizing, self-managing, self-optimizing, self-protecting) attributes are of particular value \cite{jd:componentselfstar04}\cite{zl:selfgrid04}.  This paper introduces \textit{polymorphic} self-* agents that are capable of multiple roles as directed by the environment.  These agents evolve an optimum core set of roles for which they are responsible, while still possessing the ability to take on alternate roles as environmental demands change.  They are directly implementable in computer systems applications.

The approach is based on \textit{stigmergy}, a concept that explains organization and coordination within social insect societies that rely strictly on environmental cues for indirect communication between individuals.  It is implemented on BTeV, a particle accelerator-based High Energy Physics experiment currently under development at Fermi National Accelerator Laboratory.  Multiple layers of polymorphic, very lightweight agents (VLAs) are embedded within 2500 Digital Signal Processors (DSPs) to handle fault mitigation across the system.  The primary challenge is to determine the frequency at which VLAs should perform specific monitoring tasks.  Results show how polymorphic self-* VLAs evolve independently to find the optimum rate at which monitoring and fault mitigation tasks should occur.  SWARM multi-agent simulation software is used to model RTES/BTeV.

This paper is divided into four sections.  First, some background on polymorphism and stigmergy, along with the BTeV experiment itself is provided.  A description of VLAs embedded within Level 1 of the RTES/BTeV environment is provided, followed by an explanation of current challenges and other motivating factors.  Section 3 then introduces polymorphic self-* agents and describes the design in detail.  Results of a SWARM simulation of the RTES/BTeV environment that implements the polymorphic self-* approach are then evaluated in Section 4.  Finally, next steps and a conclusion are provided.

\section{Background and Motivation}
\subsection{Polymorphism and Stigmergy}
Concepts of polymorphism and stigmergy are founded in biology and the study of self-organization within social insects.  The term \textit{polymorphism} is used in describing ants and other social biological systems, and is defined as the occurrence of different forms, stages, or types in individual organisms or in organisms of the same species, independent of sexual variations \cite{ew:evolpolyants53}\cite{jl:biochempoly65}.  Within an individual colony consisting of ants with the same basic genetic wiring, two or more castes belonging to the same sex can be found.  A caste here is defined as a differentiated morphological form with a specialized function, or at least the infrequent relict of such a form.  The function or role that any individual ant takes on is dictated by cues from the environment \cite{dw:castedetermine86}.

The agents described in detail in section 3 of this paper adhere to this definition of polymorphism in that they are genetically identical, yet each evolve distinct roles that they play as demanded of them through changes in the environment.

The concept of polymorphic agents presented in this paper is different from other definitions of polymorphism that have surfaced in computer science.  In object-oriented programming, polymorphism is usually associated with the ability of objects to override inherited class method implementations \cite{nj:oopoly02}.  The term has also arisen in other subareas of computer science, including some agent designs \cite{bb:polyeagents00}, but generally describes a templating based system or similar variation of the object-oriented model.  On the other hand, techniques that attempt to evolve specialized agents are one of the central themes under investigation in the field of large-scale multi-agent systems \cite{ps:approachlargecoord04}. 

\textit{Stigmergy} was introduced by biologist Pierre-Paul Grasse to describe indirect communication that takes place between individuals in social insect societies \cite{pg:stigmergy59}.  The theory explains how organization and coordination of the building of termite nests is mainly controlled by the nest itself, and not the individual termite workers involved.  It views the process of emergent cooperation as a result of participants altering the environment and reacting to the environment as they pass through it.  The canonical example of stigmergy is ants leaving pheromones in ways that help them find the shortest, safest distance to food or to build nests.  Ant colony optimization methods alone have had a wide impact on coordination within multi-agent systems, addressing various adaptive network routing and load balancing problems \cite{gd:antcolonyrouting98}\cite{md:antcolonyopt04}.

A stigmergic approach to fault mitigation is introduced in this paper.  Individual agents communicate indirectly through errors that they find (or do not find) in the environment.  This indirect communication is manifested through actions that each agent takes as cued by the environment.  Results show how the local actions of agents allow self-* global behavior to emerge.

\subsection{RTES/BTeV}
BTeV is a proposed particle accelerator-based HEP experiment currently under development at Fermi National Accelerator Laboratory.  The goal is to study charge-parity violation, mixing, and rare decays of particles known as beauty and charm hadrons, in order to learn more about matter-antimatter asymmetries that exist in the universe today \cite{sk:pixeldetect02}.

\begin{figure*}[t]
\centering
\includegraphics[width=130mm]{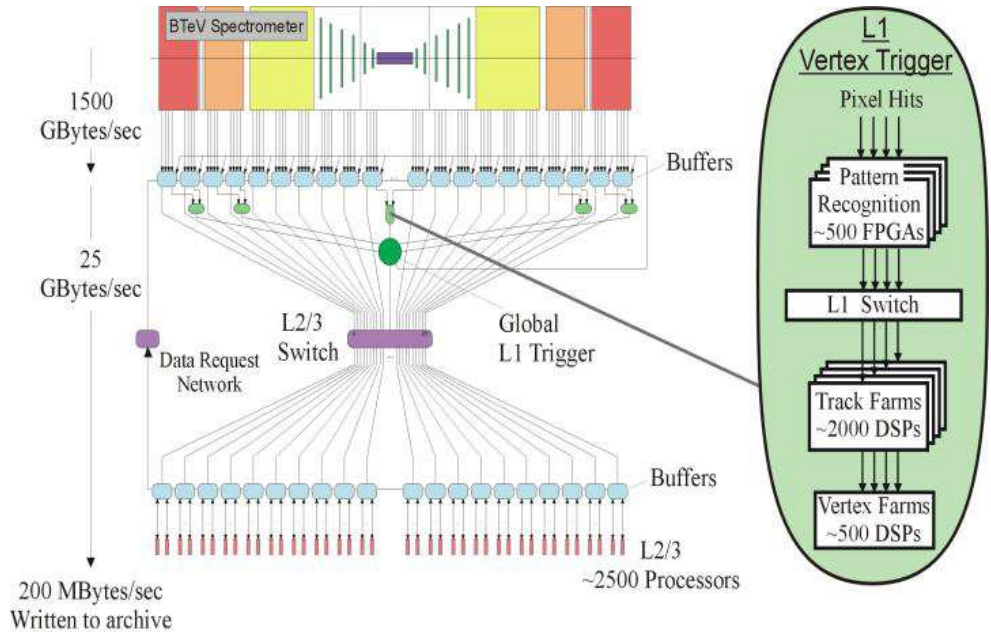}
\vspace{-.5in}
\caption{The BTeV triggering and data acquisition system showing (left side) detector, buffer memories, L1, L2, L3 clusters and their interconnects and (right side) a magnified figure of the L1 Vertex trigger.}
\label{fig:btevtrigdata}
\end{figure*}

The experiment uses approximately 30 planar silicon pixel detectors that are connected to specialized field-programmable gate arrays (FPGAs).  The FPGAs are connected to approximately 2500 digital signal processors (DSPs) that filter incoming data at the extremely high rate of approximately 1.5 Terabytes per second from a total of 20x10$^{6}$ data channels.  A three tier hierarchical trigger architecture will be used to handle this high rate \cite{sk:pixeldetect02}.  An overview of the BTeV triggering and data acquisition system is shown in Figure \ref{fig:btevtrigdata}, including a magnified view of the L1 Vertex Trigger responsible for Level 1 filtering consisting of 2500 Worker nodes (2000 Track Farms and 500 Vertex Farms).

There are many Worker level tasks that the Farmlet VLA (FVLA) is responsible for monitoring.  A traditional  hierarchical approach would assign one (or more) distinct DSPs the role of the FVLA, with the responsibility of monitoring the state of other Worker DSPs on the node \cite{fc:abstracttolerance94}.  However, this leaves the system with only very few possible points of failure before critical tasks are left unattended.

Another approach would be to assign a single redundant DSP (or more) to each and every Worker DSP, to act as the FVLA \cite{wh:sysfaulttol92}.  However, since 2500 Worker DSPs are projected, this would prove expensive and may still not fully protect all DSPs given even a low number of system failures.

The events that pass the full set of physics algorithm filters occur infrequently, and the cost of operating this environment is high.  The extremely large streams of data resulting from the BTeV environment must be processed real-time with highly resilient adaptive fault tolerant systems.

\subsection{Very Lightweight Agents (VLAs)}
Multiple levels of very lightweight agents (VLAs) \cite{jo:lightweightagents03} are one of the primary components responsible for fault mitigation across BTeV.

The primary objective of the VLA is to provide the BTeV environment with a lightweight, adaptive layer of fault mitigation.  One of the latest phases of work at Syracuse University has involved implementing embedded proactive and reactive rules to handle specific system failure scenarios.  

A scaled prototype of the Level 1 RTES/BTeV environment was presented at the SuperComputing 2003 (SC2003) conference \cite{dm:scproto04}.  Reactive and proactive VLA rules were integrated within this Level 1 prototype and served a primary role in demonstrating the embedded fault tolerant capabilities of the system.

\subsection{Challenges}

While the SC2003 prototype was effective for demonstrating the real-time fault mitigation capabilities of VLAs on limited hardware utilizing 16 DSPs, one of the major challenges is to find out how the behavior of the various levels of VLAs will scale when implemented across the 2500 DSPs projected for BTeV \cite{jk:hardwarefailure03}.  In particular, how frequently should these monitoring tasks be performed to optimize processing time, and what affect does this have on other components and the overall behavior of a large-scale real-time embedded system such as BTeV.

Given the number of components and countless fault scenarios involved, it is infeasible to design an `expert system' that applies mitigative actions triggered from a central processing unit acting on rules capturing every possible system state.  Instead, a distributed approach using self-organizing VLAs accomplishes fault mitigation within the large-scale real-time RTES/BTeV environment.

\subsection{SWARM}
SWARM (http://www.swarm.org), distributed under the GNU General Public License, is software available as a Java or Objective-C development kit that allows for the multi-agent simulation of complex systems \cite{rb:swarmsched97}\cite{md:openframework00}.  It consists of a set of libraries that facilitate implementation of agent-based models.  SWARM has previously been used by the RTES team in simulations that model the RTES/BTeV environment \cite{dm:swarmsubsum04}.

\section{Polymorphic Self-* Agents}
\subsection{Overview}
This paper introduces a stigmergic multi-agent systems approach that uses polymorphic self-* agents to address the weaknesses inherent in traditional hierarchical fault mitigation designs.  Rather than hard-wiring the assignment of FVLA roles to specific VLAs embedded within individual DSPs, VLAs are made \textit{polymorphic} so that \textit{every} VLA is equipped to play the role of FVLA for \textit{any} DSP on the same node.

Since the FVLA is responsible for a wide range of monitoring tasks, this means that we must build the capability of performing each task into every Worker Level VLA.  The classic problem this presents in traditional hierarchical approaches is how to process all of the data necessary for all of these tasks in time for a useful response \cite{mw:agenttheorypractice94}.  However, since these agents are polymorphic and evolve roles gradually over time, there is only a small set of tasks for which each agent is responsible for at any given point in time.

Stigmergy is used to determine which set of tasks any given VLA performs.  Errors found (or not found) in the environment by an individual VLA increase (decrease) the sensitivity of that VLA to that particular type of error.  Agents start out by monitoring each type of error at a fixed rate.  Then, based entirely on what is encountered in the environment, each develops a core set of roles for which it takes responsibility.  For example, a single VLA embedded within a DSP monitors each particular error at some unique rate.  When an individual VLA performs a monitoring task on some DSP, it either finds an error and performs mitigative action, or does not find an error and does nothing.  If it finds an error, it increases its own sensitivity to that type of error on the corresponding DSP.  If it does not find an error, its sensitivity to the error decreases slightly.  Results show how, over time, this produces an optimal distribution of monitoring tasks across all VLAs, with each VLA evolving responsibility for a unique core set of monitoring tasks.

The overall emergent behavior of this design results in self-organization of FVLA responsibilities based on the state and workload of all DSPs within the node.  A certain set of VLAs may perform specific FVLA tasks at one moment, and another set (which may or may not include VLAs from the original set) can be found performing these same tasks later in time.  The organization occurs automatically within the system as environmental cues fluctuate.  This eliminates the financial and efficiency costs associated with having specialized FVLAs that at times sit idle as Worker DSPs operate at full capacity and fall behind on event processing.  It also increases the efficiency of Worker DSPs that may be wasting idle time when crossing processing rates are low.  In effect, a fully connected network of FVLAs is created that continue to provide effective fault mitigation when exposed to a high volume of system failures.  

There are two key characteristics of this model.  The first is that it requires no central management or global processing.  Second, it is optimally reliable since FVLA monitoring tasks are distributed across all DSPs, and can be adapted based on changes in the environment.  The next section explains implementation details on how each individual agent uses only cues from the environment to determine necessary actions.

\subsection{Implementation}
As described above, distributed VLAs within Worker level DSPs are used to accomplish the fault monitoring tasks that the FVLA is responsible for.  However, these are the same DSPs that are responsible for the critical overall objective of Level 1 physics application (PA) data filtering \cite{sk:pixeldetect02}.  It is therefore extremely important that DSP usage by each Worker VLA is minimal, and only occurs either when the PA is not fully utilizing the DSP, or when critical fault mitigative action is required. 

Game theory has been applied to a wide range of problems, and is used here to coordinate the amount of DSP clock cycle that is allocated between the PA and the VLA.  Both the PA and VLA wish to maximize the number of clock cycles during which they have control.  If the VLA takes too many DSP cycles, then the PA will be unable to process the incoming data at a high enough rate to prevent the buffers from overflowing, resulting in a loss of data continuity.  This is often fatal for the experiment since this lost data could very well contain portions of vital characteristics of the physics properties being evaluated.  If on the other hand, the PA takes too many DSP cycles, then it runs the risk that system faults will go undetected, resulting in acceptance of corrupt data, and/or incremental bottlenecks that again cause buffer overflows. 

An efficient adaptive scheduling algorithm is required that will effectively establish scheduling priorities between the PA and VLA.  Mandatory costs associated with the Kernel/Command Processor, including clock cycle costs for context switching must be factored in.  An analysis of the worst-case behavior of tasks (both VLA and PA) can be done to determine the amount of time that must be allotted to each process.  However, there must be a way for the system to adaptively modify these values when environmental conditions change.  That is, if during every interval T, the HEP applications and the operating system use T$_{PA}$ and T$_{OS}$ time units, respectively, then the VLA will be allowed to use T -- T$_{PA}$ -- T$_{OS}$ every T time units \cite{jo:lightweightagents03}.

An analysis of best-case behavior of tasks (VLA and PA) requires the use of a \textit{utility value} in order for each DSP to determine locally precisely when the PA or VLA should relinquish control \cite{ar:gametheory00}.  A reward system based on a combination of the amount of data processed, along with the frequency of VLA maintenance checks, is used by each DSP for each error in calculating the following local utility value :\\

\noindent
DSP Utility Value = Dw$^{-1}$ + cF$^{-1}$ \hspace{.6in} , where\\

\noindent
D = Expected amount of data that DSP could process\\
\hspace*{.25in}during a given time interval (T).

\noindent
w = Current data buffer watermark.

\noindent
F = Total number of clock cycles elapsed since last\\
\hspace*{.25in}FVLA check on neighboring DSPs.

\noindent
c = Adaptive constant representing weight to place on\\
\hspace*{.25in}FVLA checks.\\

Since the amount of data that any single DSP can process (D) over a given time interval is mostly fixed, the utility value essentially involves summing the inverse of the current data buffer watermark (w$^{-1}$) with a weighted value for the inverse of the time elapsed since individual FVLA tasks were last performed (F$^{-1}$).  

The task currently active (PA or VLA) calculates the optimum expected utility value for the DSP at a time interval based on the criticality of each error.  If the active process determines that a higher DSP utility value is received by remaining active, then the active task will continue.  However, if a higher utility value can be gained by passing control to the currently inactive process, then that is what does.  For example, if the PA is currently active, the input data buffer for a given DSP is low, and FVLA monitoring responsibilities for a specific error have not been performed on a particular DSP in a long time, then the VLA task will be made active.  If however, the VLA was currently active under these conditions, then the VLA would simply maintain control for another T time steps, at which time corresponding utility values would again be calculated.  This is equivalent to determining :\\

\vspace{-.05in}
max(w, 2 $\times$ ((1 / (1 + e$^{-dF}$)) - .5)\\
\vspace{-.05in}

\noindent
the maximum value of either \textit{w} or 2 $\times$ ((sigmoid function value for \textit{F}) - .5).  Here, 2 $\times$ ((1 / (1 +e$^{-dF}$)) - .5) is an adjusted sigmoid function for F which represent \textit{F} as a weighted value between 0 and 1.  

It is important to note here that the value assigned to \textit{d} determines the steepness of the sigmoid function, and hence the sensitivity of the agent to a given error.  In other words, the higher the value of \textit{d}, the higher the adjusted sigmoid value of F, and the higher the sensitivity (the frequency of checks) of the VLA to a particular error.

\begin{figure*}[t]
\includegraphics[width=6.9in]{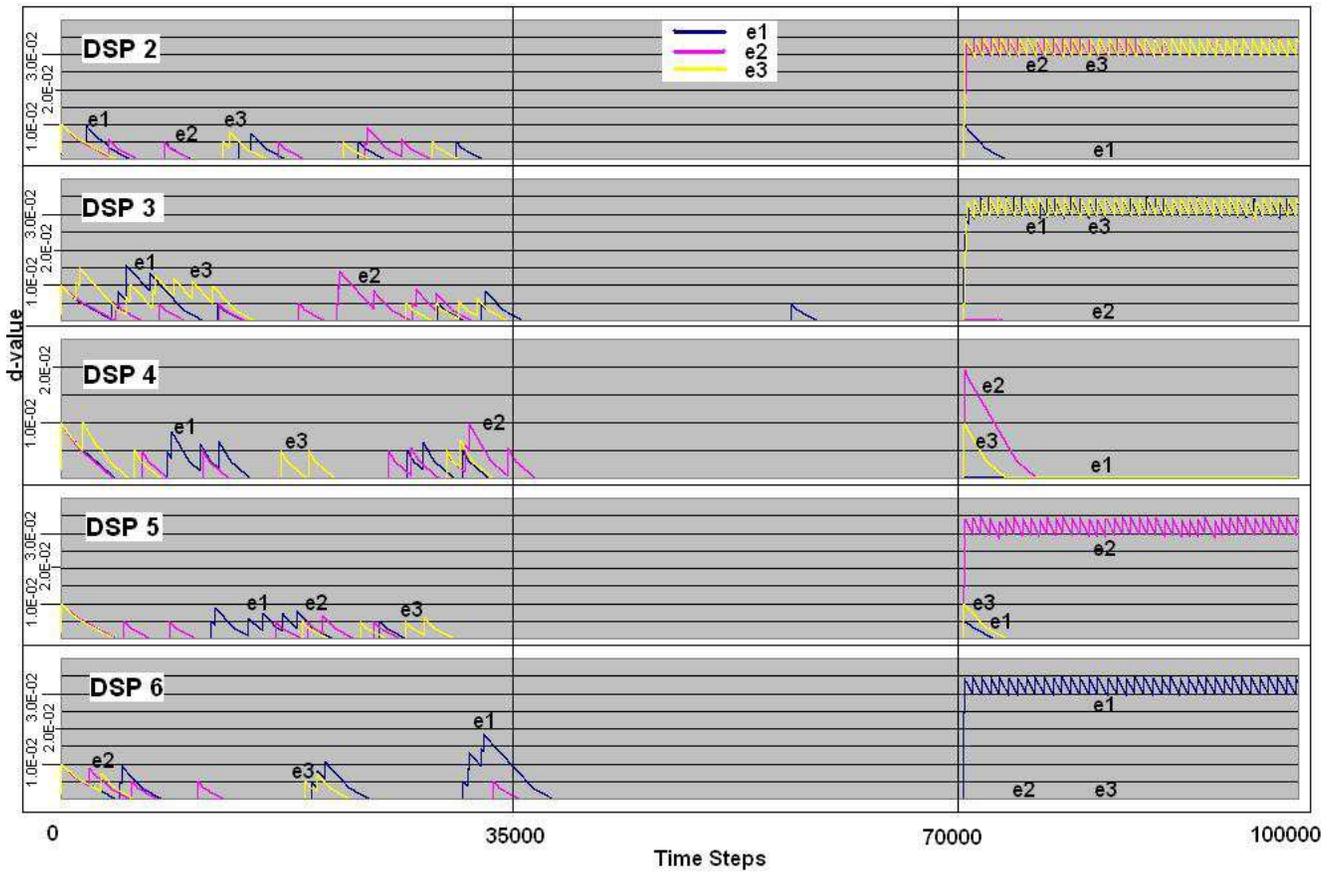}
\caption{The VLA d-value (sensitivity) for 3 distinct error types (e1, e2, e3) being monitored on DSP1.  Each of the 5 graphs represent the d-value adapted over time by each of the remaining 5 DSPs (DSP2 - DSP6) on the same Farmlet.  The simulation fluctuated the error rate between a moderate rate (5 x 10$^{-4}$) for the first 35000 time steps, a low rate (5 x 10$^{-6}$) for the next 35000 time steps (35001 - 70000), and a high rate (5 x 10$^{-3}$) for the last 30000 time steps (70000 - 100000).}
\label{fig:dvstime}
\end{figure*}

This is where the polymorphic behavior of the VLA is introduced.  Any time that an individual VLA finds a specific error while performing FVLA monitoring tasks, the \textit{d} value for that error on that particular node is increased.  Any time that an individual VLA performs a monitoring task and does \textit{not} find an error, the \textit{d} value is slightly decreased.

A high value for F means that FVLA tasks are performed more frequently (high sensitivity), whereas a low value for F means they are performed less often (low sensitivity).  The PA is passed (or maintains) control if \textit{w} is higher than this adjusted sigmoid function value for \textit{F}, otherwise the VLA is passed (maintains) control.  For example, if the PA is currently active, the input data buffer watermark for a given DSP is about half full (w=.5), and FVLA functions have recently been performed (the adjusted sigmoid function value for \textit{F} is, say, .15) then the PA will remain active.

\section{Results}
SWARM simulates Farmlet data buffer queues that are populated at a rate consistent with the behavior of the incoming physics crossing data.  Each DSP within a given Farmlet processes a fixed amount of data at each discrete time step.  Three distinct types of errors are introduced randomly within each Worker DSP at a variable rate using a Multiply With Carry (RWC8gen) random number generator with a fixed seed.  Any time a software or hardware error is encountered within the simulation, the processing rate for that DSP decreases a set amount depending on the type of  error.  The error is cleared when any DSP within the same Farmlet performs FVLA checks against the DSP for the error type present.  However, there is a time cost associated with performing these checks.  As detailed in the section above describing the self-organizing model, the DSP must decide whether or not it is worth taking time to perform FVLA monitoring tasks against neighboring DSPs.  If checks are performed too frequently, then the time available for data crossing processing is limited.  On the other hand, if they are not performed frequently enough, then the chance that other DSPs within the same Farmlet are experiencing errors is high.  As described, a high error rate will also lead to slow processing rates.  

The formula designed for these experiments calculates the frequency of performing FVLA tasks for neighboring DSPs as a sigmoid function adjusted to a value between 0.0 and 1.0.  The fullness of the crossing data buffer queue is also a value between 0.0 and 1.0 representing the data watermark percentage.  These two values are weighed against each other, and the DSP makes a decision on where to devote its energy as described in detail in the last section.

The decision of whether the VLA or PA has control of the DSP is made by each DSP at each time step in the SWARM simulation.  In this way, the monitoring tasks required by the environment are always met, but not necessarily by one (or a few) designated DSPs.  Instead, these tasks are performed by any polymorphic DSP within the Farmlet as dictated by the changing needs of the environment.  

The DSPs themselves self-organize as different DSPs within the Farmlet take on the necessary monitoring tasks at different points in time as required by the environment.  If a DSP performs FVLA monitoring tasks for a given type of error on a neighboring DSP, it will either determine that the error is not present, or it will find the error and perform the designated mitigative actions.  In the case where an error is found, the d-value for that particular error on the specific DSP is increased.  As described in detail earlier, this essentially increases the sensitivity of the VLA for this type of error.  On the other hand, if no error is found, then the d-value (sensitivity) is slightly decreased.

As detailed next, Figure \ref{fig:dvstime} shows how the local action performed by each VLA over a short period of time results in VLAs evolving responsibility for a core set of fault monitoring tasks.  Over the 100000 time steps for which the SWARM simulation is run, the 5 VLAs (1 per DSP) can be seen taking on distinct roles that lead to an efficient global fault mitigation strategy for monitoring errors on DSP1.  These roles are evolved using local information only, and rely on stigmergy within the environment for indirect coordination with other VLAs.

\begin{figure}[t]
\includegraphics[width=3.3in]{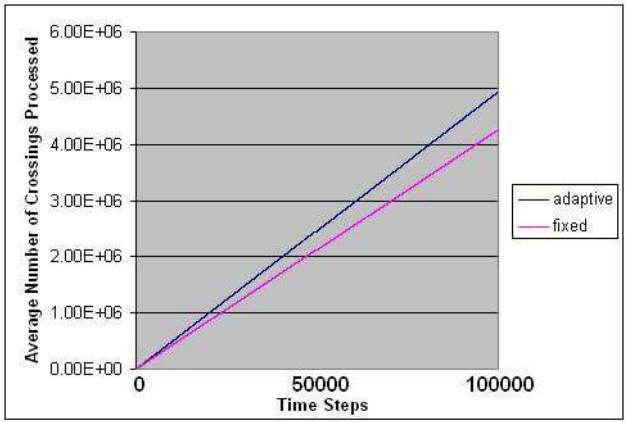}
\caption{Average number of crossings processed per DSP resulting from the stigmergic approach using polymorphic agents(adaptive), compared against the same simulation using a fixed monitoring rate (d-value fixed at .01).}
\label{fig:totalproc}
\end{figure}

The simulation fluctuates the error rate at various intervals in order to demonstrate the affect changes in error rate can have on polymorphic behavior.  A moderate error rate (5 x 10$^{-4}$) is used for the first 35000 time steps, a low error rate (5 x 10 $^{-6}$) for the next 35000 time steps (35001-70000), and the last 30000 time steps (70001-100000) use a high rate (5 x 10 $^{-3}$).  Figure \ref{fig:dvstime} shows how all of the VLAs are able to adjust sensitivity to errors on DSP1 based on these fluctuating error rates over time.  For example, the d-value (sensitivity) to individual errors on DSP1 for all 5 VLAs (embedded within DSP2 - DSP6) can be seen dropping beginning around time step 35000, and then increasing dramatically again at time step 70000 in reaction to the significant increase in error rate.

Polymorphism is demonstrated clearly in Figure \ref{fig:dvstime} which displays the VLA d-value (sensitivity) for 3 distinct error types being monitored on DSP1 within a single Farmlet.  The d-values evolved by each of the VLAs within the 5 DSPs (DSP2-DSP6) monitoring DSP1 within the same Farmlet are shown.  When the error rate is high (from time steps 70000-100000), the VLAs embedded within DSP3 and DSP6 develop a high sensitivity for error type 1 (e1), while the sensitivity for e1 of the VLAs in the remaining DSPs remains low.  Similarly, the VLAs on DSP2 and DSP5 have a high sensitivity for error type 2 (e2), and VLAs for DSP2 and DSP3 are highly sensitive to e3.  

The moderate error rate used for the first 35000 time steps reveals additional polymorphic characteristics of this approach.  Here, the error rate is not quite high enough for any single VLA to evolve long term responsibility for an individual error type on DSP1.  Instead, 1 or 2 VLAs can be seen monitoring a single error type at one moment, and then a separate VLA (or group of VLAs) can be seen monitoring the same error type a short time later.  This is due to the fact that the error rate is too low to stimulate high sensitivity in a single VLA.  Sensitivity for the error type drops to a level comparable with other available VLAs on the Farmlet.  For example, the VLAs on DSP 3 and DSP 4 develop a modest level of sensitivity for e1 early on (time steps 0-15000), but the role is taken over by VLAs on DSP 5 (time steps 15000-28000) and later DSP6 (28000-35000).

Figure \ref{fig:totalproc} shows the average data processing rate per DSP for the stigmergic approach using polymorphic agents, as compared to the same simulation using a fixed monitoring rate (d-value fixed at .01) for each agent.  The polymorphic agents in the stigmergic approach adapt an optimum monitoring rate for each error based strictly on the demands of the environment at any given time.  This results in a higher number of crossings processed since, as described in detail earlier, less time is wasted performing needless monitoring tasks or missing critical errors.

\section{Next Steps}
The next phase of this project will expand the number of different types of errors handled, along with the amount of fluctuation in error rates.  It will also focus further on how sensitivity (d-value) is adapted for each VLA.  Currently, a rudimentary method is used that slightly increases (or decreases) sensitivity based on the presence (or absense) of an error.  Other variables could be considered in determining the amount of change to apply, such as factoring in the severity level of the error, or looking at the consequences of other recently taken actions.  An enhanced evaluation methodology to better demonstrate the performance advantage of this approach as compared to other traditional methodologies is also necessary.

Another issue being investigated is how to handle communication between agents when one agent has information that may be relevant to other agents, but it does not know to which other agent the information is relevant.  This is a problem encountered in many large-scale multi-agent systems \cite{ps:approachlargecoord04}, and is especially an issue in fault mitigation where trends in information received across agents can provide valuable warning signs.

At the same time, another scaled prototype of the actual projected RTES/BTeV software and hardware environment based on the SC2003 demonstration system is also being developed, and will integrate the VLA self-* model.  This prototype will be presented at the 2nd Workshop on High-Performance Fault-Adaptive Large-Scale Embedded Real-Time Systems (FALSE-II) in the IEEE Real-Time and Embedded Technology and Applications Symposium (RTAS05).

\section{Conclusion}
This paper has described a fully distributed stigmergic approach to fault mitigation in large-scale real-time systems using lightweight, polymorphic, self-* agents embedded within individual DSPs.  Stigmergy facilitates indirect communication and coordination between agents using cues from the environment, and concepts from game theory and polymorphism allow individual agents to evolve a core set of roles for which it is responsible.  Agents adapt these roles as environmental demands change.  The approach is implemented on a SWARM simulation of BTeV, a High Energy Physics experiment consisting of 2500 DSPs.  

Results demonstrate the polymorphic nature of the agents, and display the performance and reliability advantages of this approach.  The next phase of this project will increase the number of possible error types, and add more fluctuation to individual error rates.More sophisticated ways of adapting error sensitivity among agents will also be investigated, along with more elaborate performance evaluation metrics.

\section{Acknowledgements}
The research conducted was sponsored by the National Science Foundation in conjunction with Fermi National Laboratories, under the BTeV Project, and in association with RTES, the Real-time, Embedded Systems Group.  This work has been performed under NSF grant \# ACI-0121658.

\bibliography{aamas05}

\end{document}